\documentclass[10pt,twocolumn,letterpaper]{article}

\usepackage{cvpr}              
\usepackage{adjustbox}
\usepackage[accsupp]{axessibility}
\definecolor{cvprblue}{rgb}{0.21,0.49,0.74}
\usepackage[pagebackref,breaklinks,colorlinks,allcolors=cvprblue]{hyperref}

\def\paperID{34378} 
\def\confName{CVPR}
\def\confYear{2026}

\title{CLLAP: Contrastive Learning-based LiDAR-Augmented Pretraining for Enhanced Radar-Camera Fusion}
\author{
Bingyi Liu$^{1}$ \quad
Chuanhui Zhu$^{1}$ \quad
Hongfei Xue$^{2}$\thanks{Corresponding author.} \quad
Jian Teng$^{1}$ \\
Jipeng Liu$^{5}$ \quad
Enshu Wang$^{3}$ \quad
Penglin Dai$^{4}$ \quad
Pu Wang$^{2}$ \\[0.5em]
$^{1}$Wuhan University of Technology \quad
$^{2}$University of North Carolina at Charlotte \quad
$^{3}$Wuhan University \\
$^{4}$Southwest Jiaotong University \quad
$^{5}$The Hong Kong Polytechnic University
}

\begin{document}
\maketitle
\begin{abstract}
Accurate 3D object detection is critical for autonomous driving, necessitating reliable, cost-effective sensors capable of operating in adverse weather conditions.
Camera and millimeter-wave radar fusion has emerged as a promising solution; however, these methods often rely on finely annotated radar data, which is scarce and labor-intensive to produce. 
To address this challenge, we present \textbf{CLLAP}, a \textbf{C}ontrastive \textbf{L}earning-based \textbf{L}iDAR-\textbf{A}ugmented \textbf{P}retraining framework that enhances the performance of existing radar-camera fusion methods for 3D object detection.
CLLAP leverages abundant LiDAR data to generate pseudo-radar data using the proposed L2R (LiDAR-to-Radar) Sampling method.
Then, it incorporates this data into a novel dual-stage, dual-modality contrastive learning strategy, enabling effective self-supervised learning from paired pseudo-radar and image data.
This approach facilitates effective pretraining of existing radar-camera fusion models in a plug-and-play manner, enhancing their feature extraction capabilities and improving detection accuracy and robustness.
Experimental results using NuScenes and Lyft Level 5 datasets demonstrate significant performance improvements across three baseline models, highlighting CLLAP’s effectiveness in advancing radar-camera fusion for autonomous driving applications.

\end{abstract}
\vspace{-15pt}
\section{Introduction}
\begin{figure}[!t]
\centering
\includegraphics[width=0.5\textwidth]{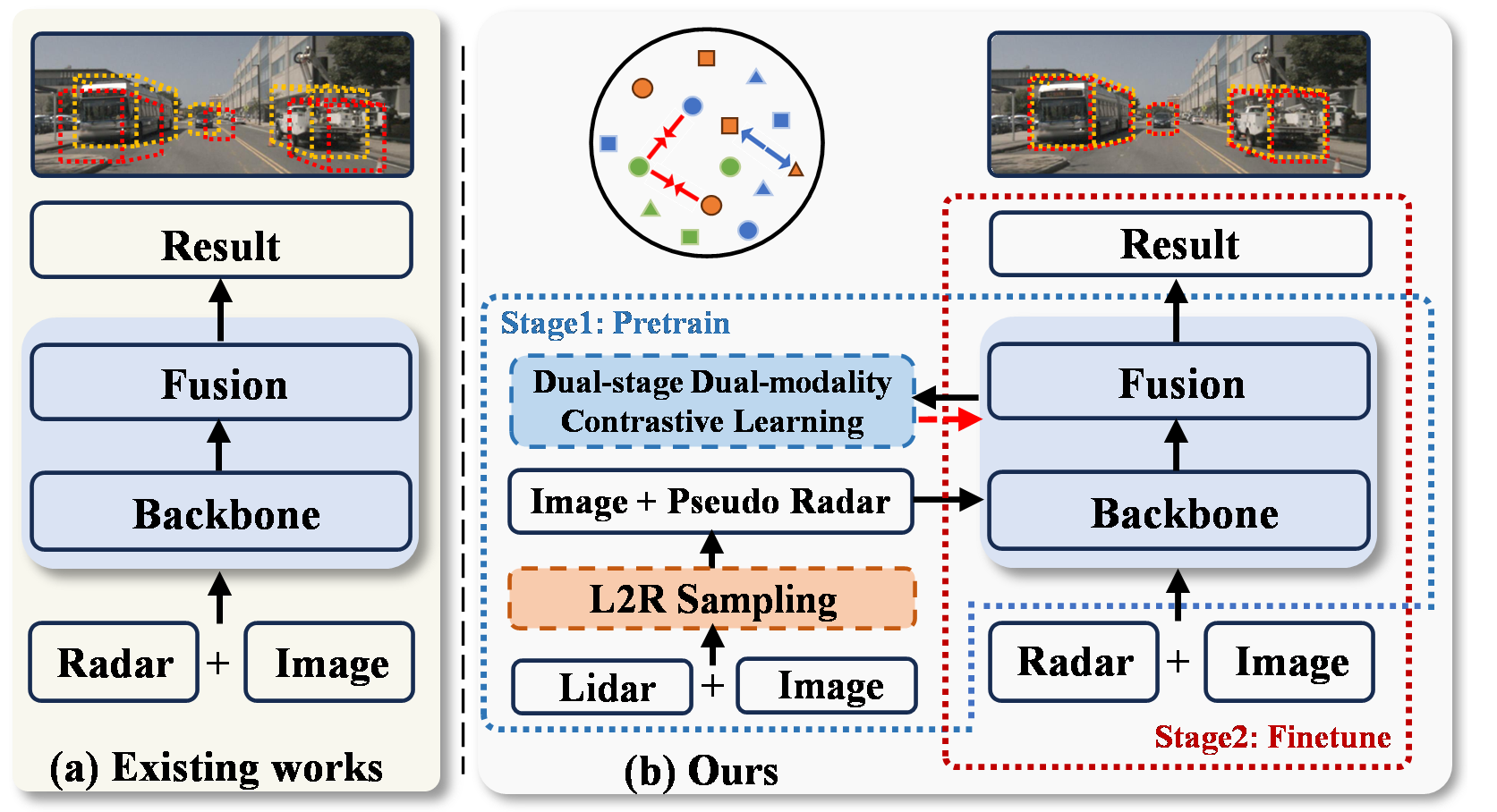}
\vspace{-18pt}
\caption{\small{(a) Existing camera-radar fusion models rely solely on scarce, finely annotated radar-image data for training, leading to suboptimal performance.
(b) The proposed CLLAP framework addresses this limitation by generating pseudo-radar data from large-scale unlabeled LiDAR data and employing a novel dual-stage, dual-modality contrastive learning strategy for self-supervised pretraining of the fusion model. Using the pretrained model as a starting point to train the fusion models in (a) significantly improves their detection performance and robustness.}}
\label{introduction}
\vspace{-22pt}
\end{figure}

3D object detection plays a crucial role in autonomous driving by enhancing spatial understanding, situational awareness, and decision-making, enabling autonomous driving systems to navigate complex environments with precision and safety \cite{peng2024vision,liu2024v2x}.
However, relying on a single sensor can limit performance, prompting the adoption of multi-modality sensor fusion. By integrating data from cameras, radar, and LiDAR, multi-modality fusion offers a richer and more robust perception of the environment, significantly improving 3D detection accuracy.
Among these approaches, camera-radar fusion is a cost-effective and resilient approach for real-world applications \cite{yao2023radar}. Unlike expensive and interference-prone LiDAR, radar offers reliable performance in adverse weather, making it ideal for intelligent transportation systems. This fusion delivers robust, affordable solutions for smarter, safer cities \cite{fan20244d}. 
CRN \cite{kim2023crn} and RCBEVDet \cite{lin2024rcbevdet} are notable radar-camera fusion frameworks. CRN utilizes multi-modality deformable attention to generate a semantically rich bird’s-eye-view representation from image features, while RCBEVDet leverages a dedicated cross-attention fusion framework to align radar and camera modalities in a unified space.

Despite recent progress, further advancement in radar-based perception remains limited by the scarcity of large, well-annotated open-source radar datasets, which are crucial for both training and evaluation \cite{barbosa2023camera}.
The intrinsic sparsity and ambiguity of radar measurements also make manual annotation difficult and labor-intensive, further slowing dataset development \cite{wang2024crrfnet}.
Although several data augmentation strategies attempt to mitigate this limitation—such as rotating or duplicating radar point clouds, or transplanting objects across datasets to increase diversity—these operations often introduce non-negligible artifacts and inconsistencies.
As a result, while such techniques may offer modest robustness gains, the errors they introduce ultimately limit their ability to address the fundamental challenge of radar data scarcity \cite{raghunathan2020understanding}.


Recent LiDAR-guided knowledge distillation methods~\cite{xu2025sckd,bang2025rctdistill,bang2024radardistill,zhao2024crkd} treat LiDAR as a privileged teacher and transfer its rich geometric cues to radar detectors through supervised distillation on paired, annotated radar–LiDAR data, thereby boosting radar-based 3D object detection. This raises a natural question: \textit{can we exploit abundant LiDAR data in a label-free manner, by converting LiDAR into pseudo-radar measurements and using them to self-supervise radar–camera fusion pretraining, so that radar perception can be enhanced without relying on large-scale annotated radar–LiDAR datasets?}

To this end, we propose CLLAP, a novel \textbf{C}ontrastive \textbf{L}earning-based \textbf{L}iDAR-\textbf{A}ugmented \textbf{P}retraining framework.
CLLAP generates pseudo-radar data from LiDAR data and incorporates a designed self-supervised contrastive learning strategy for effective model pretraining to enhance the sensing performance of existing radar-camera fusion models.
Specifically, the CLLAP framework introduces two key innovations. First, L2R (LiDAR-to-Radar) Sampling generates high-quality pseudo-radar data from LiDAR, replicating radar-specific characteristics, including an estimated velocity distribution absent in the original LiDAR data.
This method leverages large-scale, unlabeled LiDAR datasets to enrich training data diversity, enabling robust model training.
Second, Dual-Stage Dual-Modality Contrastive Learning employs contrastive learning to align features from radar and camera modalities that represent the same object or scene. By combining cross-modal local and global contrastive losses with intra-modal global contrastive loss, this strategy enables robust self-supervised learning. Pretraining on unlabeled images and pseudo-radar data enhances feature alignment and representation across modalities, leveraging their complementary strengths. Together, these innovations reduce reliance on large paired radar-camera datasets, improve feature learning, and significantly enhance the performance and robustness of radar-camera fusion models. The main contributions are summarized as follows:
\begin{itemize}
\item We introduce \textbf{CLLAP}, a novel Contrastive Learning-based LiDAR-Augmented Pretraining framework that leverages large-scale pseudo-radar data generated from high-quality, open-source LiDAR datasets for self-supervised pretraining, addressing the scarcity of radar data and enhancing the performance of existing radar-camera fusion models.

\item We develop the L2R Sampling method to generate high-quality pseudo-radar data from LiDAR and propose a Dual-Stage Dual-Modality Contrastive Learning strategy, improving the model’s representation capabilities by carefully aligning and enhancing complementary features from radar and camera modalities.
\item Extensive experiments on the NuScenes and Lyft Level 5 datasets demonstrate that our proposed pretraining framework consistently enhances the performance of three different baseline radar-camera fusion models.
\end{itemize}
\section{Related Work}

\begin{figure*}[htbp]
\centering
\includegraphics[width=0.95\textwidth]{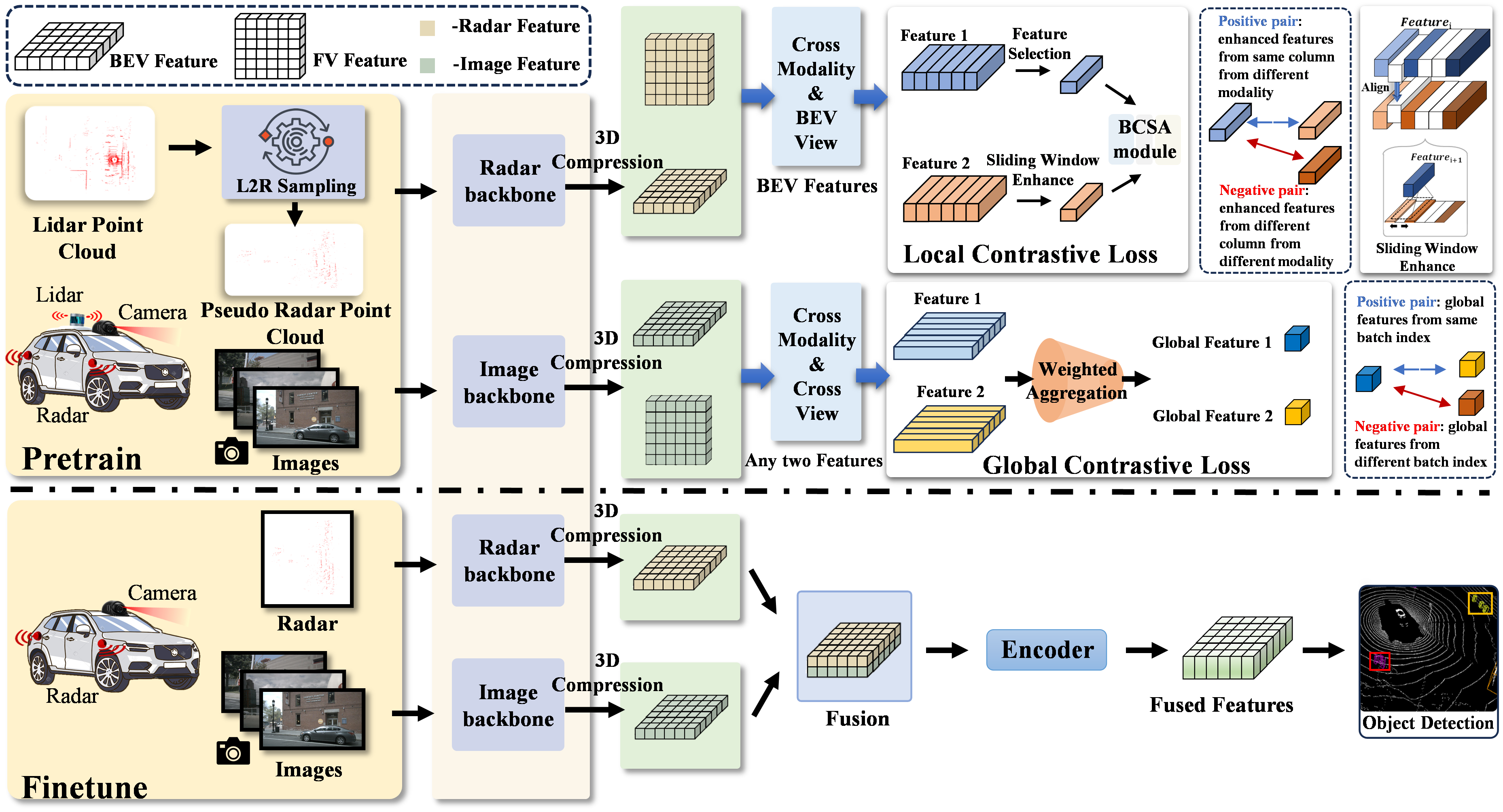}
\vspace{-10pt}
\caption{\small{The overall architecture of the proposed CLLAP Method. The framework begins by generating pseudo-radar point clouds from LiDAR data using the L2R Sampling method. Combined with image data, the backbone processes these inputs to extract features, pre-trained using the proposed dual-stage, dual-modality contrastive learning strategy. The model is then fine-tuned with real radar data for 3D object detection task.}}
\label{fig:our model}
\vspace{-15pt}
\end{figure*}

\subsection{Camera-radar fusion 3d object detection}

Accurate 3D object detection using radar is challenging due to the sparse radar point cloud and limited 3D semantic information. To overcome these limitations, fusion approaches combining radar and camera data have been explored. CRAFT \cite{kim2023craft} utilizes early fusion to integrate spatial and contextual features. ClusterFusion \cite{kurniawan2023clusterfusion} preserves local spatial features through point cloud clustering, minimizing information loss. MVFusion \cite{wu2023mvfusion} employs a Semantically Aligned Radar Encoder and a Radar-Guided Fusion Transformer for enhanced global correlation. BEV-radar \cite{zhao2024bev} integrates data bi-directionally in the BEV domain, while BEVCar \cite{schramm2024bevcar} uses attention-based enhancement initialized by sparse radar points to refine image representation. Unlike the above methods trained only on radar datasets, our approach leverages LiDAR data for pretraining, leading to improved performance.

\subsection{Data Augmentation}
To further enhance the performance of fusion-based detection, many studies have proposed data augmentation techniques to increase the diversity and robustness of training data.
PasteFusion \cite{yuan2024pastefusion} introduces the Paste-Aug algorithm, which improves image and point cloud data while addressing occlusion issues and reducing computational demands. 
Deformable Feature Fusion Network \cite{guo2024deformable} further advances 3D object detection by incorporating the Feature Alignment Transform (FAT) module, which mitigates feature misalignment in point cloud enhancement by recording transformation matrices and inverting voxel coordinates. 
C-CLOCs \cite{zhang2024contrastive} improves multi-modality 3D object detection through Multimodal Ground Truth Sampling (MGS), which automates database generation via instance segmentation. Our L2R Sampling method enhances generalization by sampling real LiDAR data matching the radar point cloud distribution, rather than generating synthetic data for augmentation.

\subsection{Contrastive Learning}
Self-supervised learning (SSL) leverages pseudo-labels to generate representations for downstream tasks, with contrastive learning emerging as a key technique in computer vision \cite{le2020contrastive}. In multi-modality 3D object detection, innovative SSL methods have advanced the field. CrossPoint \cite{afham2022crosspoint} employs intra- and cross-modality losses to improve 3D point cloud understanding. ContrastAlign \cite{song2024contrastalign} aligns heterogeneous modalities for enhanced cross-modality 3D detection. $\text{CLIX}^{\text{3D}}$ \cite{hegde2024multimodal} achieves domain-invariant feature learning with supervised contrastive learning. MODCL \cite{lan2024modcl} integrates point cloud-image mapping and supervised contrastive learning, surpassing traditional multi-modality methods on the SUN RGB-D dataset. In contrast to the aforementioned contrastive learning schemes, our method takes full advantage of the fusion models by applying multi-modality contrastive learning at both local and global stages.

\subsection{Cross-Modal Knowledge Distillation}
Cross-modal knowledge distillation (KD) leverages LiDAR as a privileged teacher to improve radar-based 3D perception. SCKD~\cite{xu2025sckd}, RadarDistill~\cite{bang2024radardistill}, CRKD~\cite{zhao2024crkd}, and RCTDistill~\cite{bang2025rctdistill} all distill rich LiDAR (or LiDAR–camera) representations into radar or radar–camera students via supervised feature and proposal alignment on paired, labeled radar–LiDAR data, consistently boosting detection performance. However, these methods inherently depend on large-scale annotated radar–LiDAR corpora and treat LiDAR only as a teacher whose predictions or intermediate features serve as distillation targets for specific radar or radar–camera samples. In contrast, our CLLAP framework does not perform supervised distillation from LiDAR outputs; instead, it treats large-scale \emph{LiDAR-only} point clouds as an unlabeled geometric corpus, converts them into pseudo-radar measurements, and pairs them with images to conduct self-supervised contrastive pretraining of radar–camera fusion models, thereby improving radar perception without paired radar–LiDAR data or detection labels.


\section{The Proposed Method}

\subsection{Overview}

\cref{fig:our model} illustrates the architecture of our proposed unsupervised contrastive learning method CLLAP, comprising two stages: pre-training and fine-tuning, divided into three steps.
In the first step, the L2R Sampling module generates pseudo-radar data from large-scale LiDAR data, enabling initial pre-training of the feature encoder to capture diverse feature representations. The second step employs real radar data for secondary pre-training, enhancing the model's understanding of real-world radar data distribution. Together, these steps form the pre-training stage, leveraging a dual-stage, dual-modality contrastive learning approach to capture intrinsic correlations across modalities and views. In the fine-tuning stage, the third step uses real radar data for task-specific optimization. Section \ref{L2R} details the L2R Sampling module, while Section \ref{Dual-Stage Dual-Modal Contrastive Learning Strategy} discusses the Dual-Stage Dual-Modality Contrastive Learning Strategy.

\subsection{L2R Sampling module}
\label{L2R}

\begin{figure*}[!t]
\centering
\includegraphics[width=0.95\textwidth]{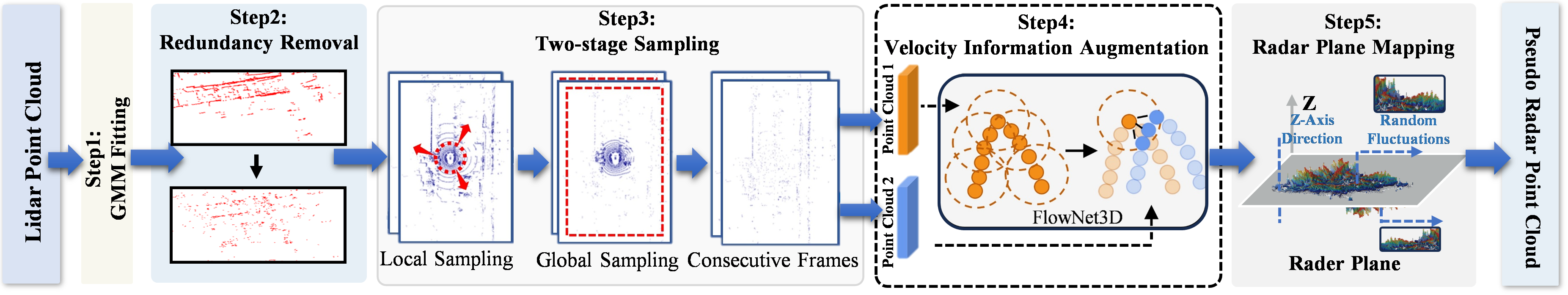}
\vspace{-12pt}
\caption{\small{Pipeline of the L2R Sampling process for generating pseudo-radar point clouds from LiDAR data. }}
\label{L2R Sampling module ways}
\vspace{-18pt}
\end{figure*}

In this section, we introduce the design of our L2R Sampling module, illustrated in \cref{L2R Sampling module ways}. Unlike prior neural network-based methods for converting LiDAR to radar data \cite{lee2024lidar,wang2020l2r}, our approach emulates radar point cloud distribution characteristics without training neural networks. This method bridges the distribution gap between LiDAR and radar data, enabling the effective use of abundant LiDAR datasets for radar-related tasks. The pseudo-radar data generation process comprises five steps: Gaussian Mixture Model (GMM) fitting, point cloud redundancy removal, two-stage sampling, velocity augmentation, and radar plane mapping.

The first step in sampling is GMM fitting. Radar point clouds in traffic scenarios exhibit highly non-uniform densities—sparse in most frames with occasional dense clusters. To realistically capture these patterns without relying on artificially uniform data, we model the density distribution using a GMM. Experiments (Appendix) show that variations in GMM parameters have minimal impact on our framework.

The second step is to use KD-tree-based spatial indexing method to remove redundancy. For each point, we calculate its minimum distance to neighboring points $d_{\text min}$, removing all points whose distance to other points is below a threshold $d_\text {threshold}$. The resulting point cloud, after redundancy removal, is sparser, preventing the concentration of points in any specific region, and better simulates the characteristics of radar point clouds.

In order to effectively simulate the sparse distribution characteristics of the radar point cloud in space, we perform the third step: two-stage sampling. LiDAR point clouds are typically denser in the center region, whereas radar point clouds exhibit a more uniform distribution across the entire space. To reflect this, we apply a weight-based sampling method to sample point clouds from both the non-central region and the global space. 
Based on the point cloud count $N$ derived from the GMM fitting, we divide it equally into two sampling categories: the non-central region sample count $N_1$ and the global region sample count $N_2$, where $N = N_1 + N_2$.
First, we sample $N_1$ point clouds located farther than a certain threshold (set to 15 meters in our experiments) from the center to obtain the external sampled point clouds. Then, the remaining points $N_2$ are sampled from the global region. This approach ensures that the sampled point clouds more accurately reflect the distribution characteristics of radar point clouds.
During sampling, we incorporate point cloud intensity, sparsity, and distance to compute a weighted sampling weight for each point.
The intensity weight $w_{\mathrm{int}}$ reflects the relative contribution of each point based on its point intensity. To ensure points with greater reflective intensity are properly emphasized during sampling, greater weights are assigned to them. The intensity weight for each point is calculated as $w_{\mathrm{int}} = I_i^{\frac{1}{2}} / \sum_j I_j^{\frac{1}{2}}$, where $I_i$ is the intensity of point $i$. 
To ensure accurate feature representation, we assign weights based on point sparsity and distance. Sparse regions require greater sampling to capture their structural and semantic information, quantified by the sparsity weight $w_{\mathrm{spa}}={\sum_jD_{ij}^2}$, where $D_{ij}$ is the Euclidean distance between point $i$ and its $j$-th nearest neighbor. Similarly, points farther from the center of the point cloud, often sparse in radar sensing, are assigned a distance weight  $w_{\mathrm{dist}}={1} /{D_{iO}^2}$, where $D_{iO}$ is the Euclidean distance from point $i$ to the origin. 
Finally, the overall sampling weight is a linear combination of the three individual weights, with scaling factors $\alpha_{int}$, $\alpha_{spa}$, $\alpha_{dist}$ controlling the contributions of each weight:
$w_{\mathrm{final}}=\alpha_{\mathrm{int}}w_{\mathrm{int}}+\alpha_{\mathrm{dist}}w_{\mathrm{dist}}+\alpha_{\mathrm{spa}}w_{\mathrm{spa}}$, where $\alpha_{int} = 4$, $\alpha_{spa} = 2$,  $\alpha_{dist} = 4$ in our experiment (appendix). By normalizing $w_{\mathrm {final}}$ for all points, we ensure the sampling probabilities are well-defined. 

To supplement the missing velocity information in the LiDAR data and align the motion between frames, in the fourth step, we use FlowNet3D \cite{liu2019flownet3d} to predict the scene flow between consecutive frames, completing the velocity information augmentation step. Using the FlowNet3D model, we can predict velocity information for each LiDAR point, further enhancing the motion characteristics of the generated pseudo-radar point clouds. 

\begin{figure}[!t]
\centering
\includegraphics[width=0.48\textwidth]{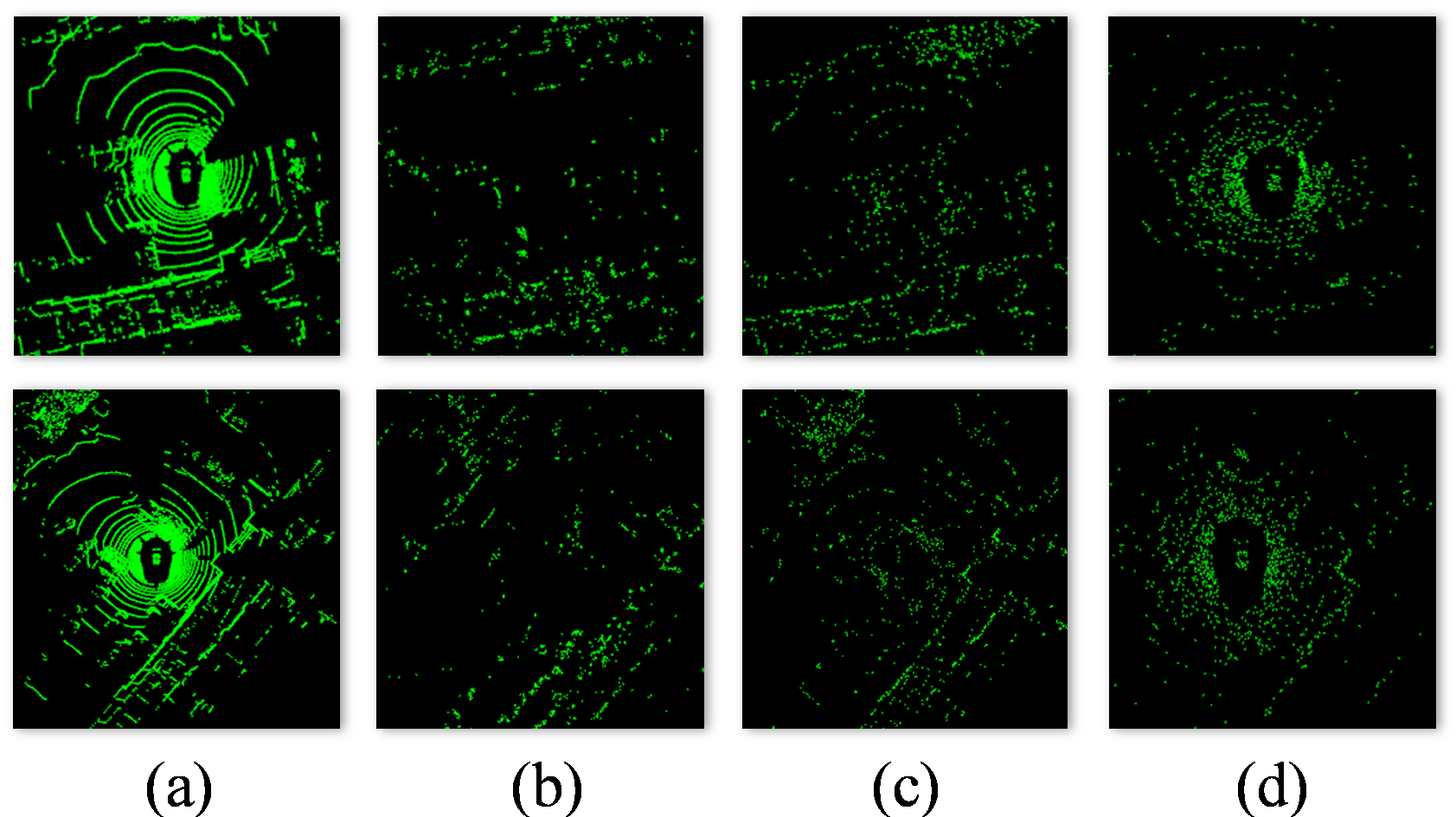}
\vspace{-20pt}
\caption{\small{Comparison of sampling methods visualization. This figure illustrates different point cloud sampling methods: (a) represents the original LiDAR point cloud frames, (b) shows the ground-truth radar point cloud frames, (c) displays the pseudo-radar point cloud frames generated by the L2R Sampling module, and (d) depicts the point cloud frames sampled with distance as the weighting factor.}}
\label{L2R view}

\end{figure}

Since radar usually does not provide accurate altitude information, we map the point cloud to the radar plane in the final step of radar plane mapping.
Due to the random nature of the sampling itself, our method can generate pseudo-radar point clouds with different distributions for the same scene, thus providing diverse datasets for model training and further enhancing the generalization ability of the model. The visual results of the L2R operation are shown in the \cref{L2R view}.

\subsection{Dual-Stage Dual-Modality Contrastive Learning Strategy}
\label{Dual-Stage Dual-Modal Contrastive Learning Strategy}
Given the scarcity of open-source datasets and the challenges associated with labeling 3D radar point clouds, further improvements in model performance are limited by the available data. To address this, we propose leveraging unsupervised contrastive learning to pre-train the model using unlabeled radar data and pseudo-radar point clouds generated through the L2R Sampling module. This pre-training process aims to enhance model performance by using a contrastive learning framework that incorporates both Local Contrastive Loss and Global Contrastive Loss.

\begin{figure}[!t]
\vspace{-14pt}
\centering
\includegraphics[width=0.40\textwidth]{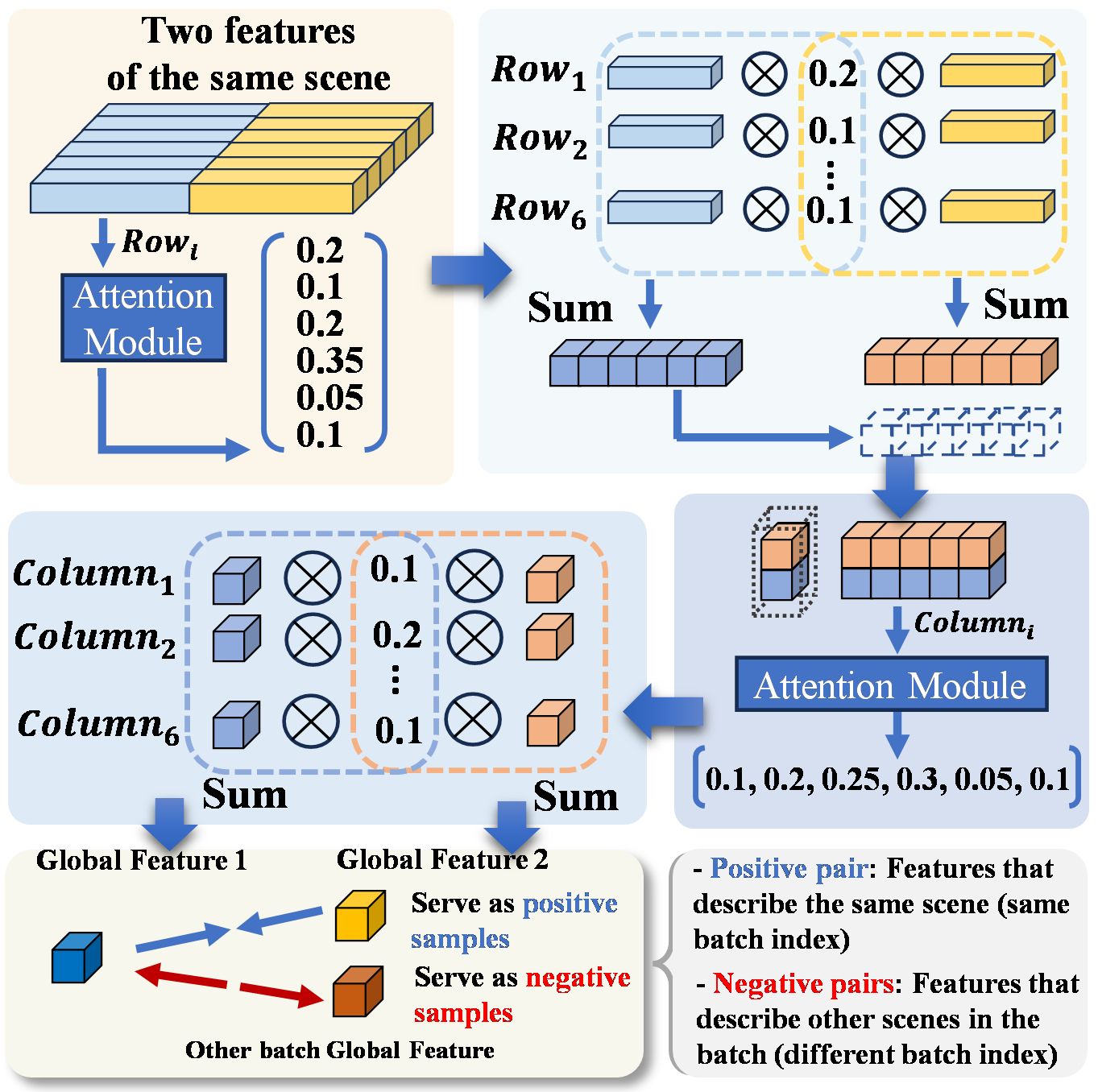}

\caption{{An overall framework for Global Contrastive Loss. The inputs consist of features from multiple views across different modalities, grouped into six sets of contrast losses. An attention mechanism is applied to perform weighted aggregation of features into global features, followed by the computation of contrast losses.}}
\label{The Global Contrastive Loss}
\vspace{-16pt}
\end{figure}

\subsubsection{Local Contrastive Loss.}
Different modalities, such as radar and image, often provide complementary information. However, even when their feature maps are structurally aligned—such as both being represented in the bird's eye view (BEV) format—significant semantic gaps can emerge due to inherent differences in their sensory characteristics. 

To address this issue, we align corresponding feature columns across modalities to substantially reduce the semantic gap and enhance the quality of fused features. By aligning column features, we can directly optimize localized feature correspondence, which is particularly crucial for tasks requiring spatial precision, such as object detection and localization. The reason for choosing a column feature for the loss function stems from the observation that corresponding columns across different modalities and views represent the same underlying features. 
This alignment is maintained across views, underscoring the importance of columns as units of correspondence for optimization. 

\begin{figure}[!t]
\centering
\includegraphics[width=0.48\textwidth]{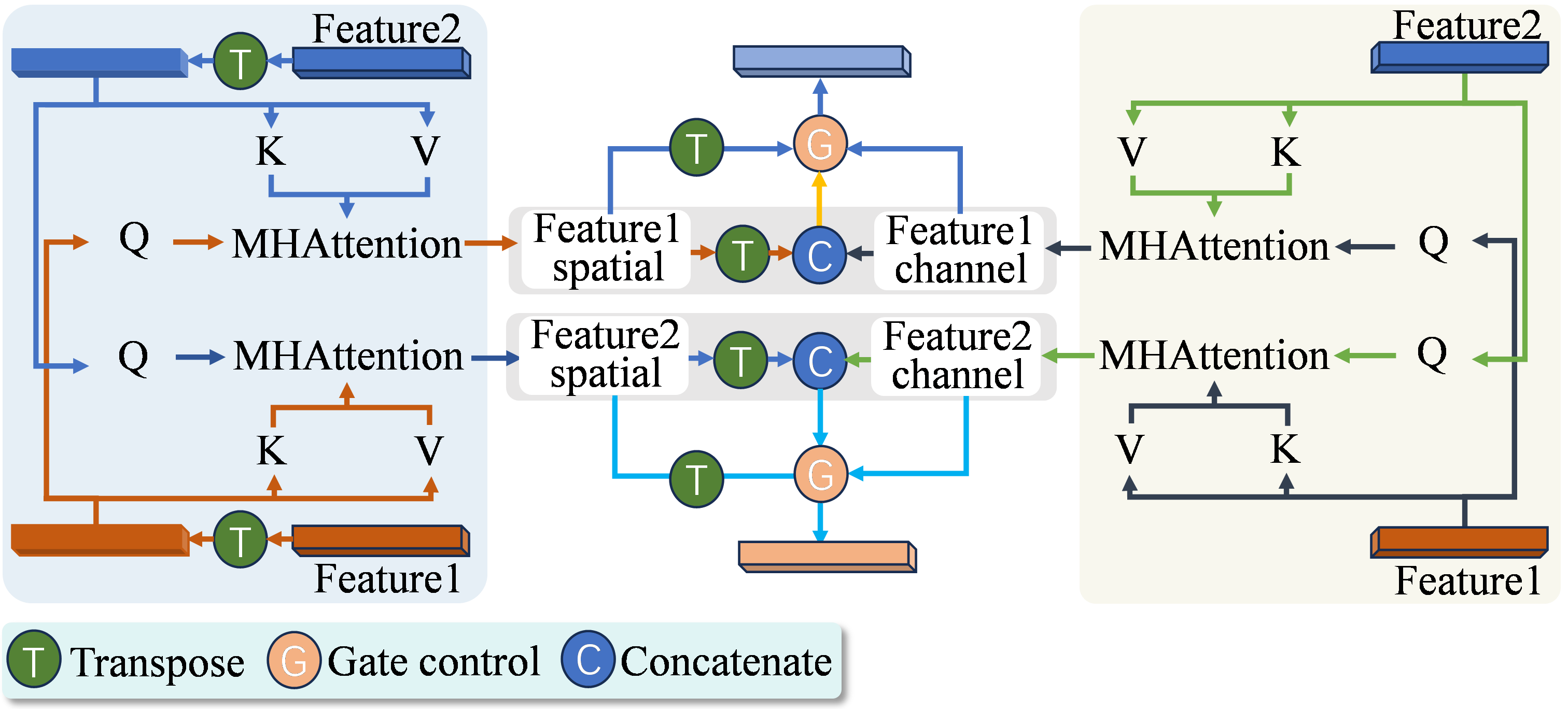}
\vspace{-15pt}
\caption{\small{The overall architecture of the proposed BCSA Module. }}
\label{BCSA}
\vspace{-15pt}
\end{figure}

Building on this insight, we propose the Local Contrastive Loss as shown in \cref{fig:our model}, which enforces alignment between feature columns across different modalities. 
To implement this, we first extract the BEV feature based on the baseline model from the radar ($F_\text {rad}$) and the camera ($F_\text {img}$) modalities. Select $N$ (number of batch size) feature columns from each modality. Each set of corresponding columns represents the same spatial position across the modalities.
For fine-grained feature misalignment, we conduct contrastive learning on carefully designed positive sample pairs.
Specifically, for each feature column, we define a local search area with a width of $R(R = 5)$ in the query feature map and use a sliding window of width $r (r = 3, r\textless R )$ to slide through the local search area to extract $n=R-r+1$ candidate areas.
Each candidate area is adaptively aggregated via an attention module to match the feature column’s dimensions, followed by pairwise similarity comparison.
The highest-scoring candidate is selected as a positive pair, effectively replacing direct positional alignment with a learned, similarity-driven matching strategy.
This approach enhances feature correspondence precision while maintaining robustness to spatial discrepancies between modalities.
Furthermore, we introduce a Bidirectional Channel-Spatial Attention module, as shown in \cref{BCSA}.
For calculated positive pairs, their channel and spatial dimensions of the feature maps contain distinct semantic meanings.
This inspired us to decoupling the modeling of channel and spatial dimensions, allowing the model to conduct finer-grained feature alignment.
Specifically, our module independently applies attention mechanisms along the channel and spatial dimensions.
\textit{Channel Attention} learns importance weights for different feature channels, highlighting those most discriminative for the current task. \textit{Spatial Attention} learns importance weights for different spatial locations within the feature column, emphasizing the importance of the region of interest in object detection while suppressing interference from background regions. Through this bidirectional decoupling and weighting across channel and spatial dimensions, the module adaptively refines the most relevant feature information for positive pair contrastive learning, suppressing irrelevant noise and guiding the model to learn more discriminative feature representations.

We use InfoNCE \cite{oord2018representation} loss to minimize the distance between positive pairs and maximize the distance from negative pairs:
\vspace{-10pt}
{\footnotesize
\begin{equation}
\label{eq:lcol}
\mathcal{L}_{\text{col}} = -\frac{1}{N} \sum_{i=1}^{N} \log \left( 
    \frac{\exp \left(\operatorname{sim}\left(\mathbf{f}_{i}^{\text{rad}}, \mathbf{f}_{i}^{\text{img}}\right) / \tau\right)}
         {\sum_{j=1}^{N} \exp \left(\operatorname{sim}\left(\mathbf{f}_{i}^{\text{rad}}, \mathbf{f}_{j}^{\text{img}}\right) / \tau\right)} 
\right)
\end{equation}
}
where $\operatorname{sim}(\cdot, \cdot)$ denotes cosine similarity, and $\tau$ is a temperature parameter that controls the smoothness of the softmax distribution. This loss encourages the contents of the same columns in different modalities to be more similar and the contents of different columns to be less similar.

\subsubsection{Global Contrastive Loss.}
While Local Contrastive Loss effectively aligns corresponding features at a fine-grained level, it may lead to semantic disjunction between features across different columns. To address this, we introduce Global Contrastive Loss, as shown in \cref{The Global Contrastive Loss}, to better leverage multi-view and multi-modality representations.

In multi-modality feature fusion, the routine is to exploit complementary information from diverse sensory modalities (e.g., radar and camera). However, relying solely on simple modality-based fusion often overlooks the important variations that arise from multiple viewpoints. Each view offers unique insights that are crucial for forming a more complete and accurate representation of objects.
By combining these views, we can enhance the model's ability to understand the scene from multiple dimensions, improving model robustness and accuracy. Therefore, we propose a novel approach that integrates both multi-modality and multi-view features, effectively combining their respective advantages to optimize feature extraction and fusion. This method not only bridges the semantic gaps between modalities but also ensures the model can capture the variations across different views, resulting in a more comprehensive and effective representation for downstream tasks.

Considering all views and modalities, our approach results in six sets of contrastive losses: 
$(F_{\text img}^{\text bev}, F_{\text img}^{\text fv})$,
$ (F_{\text img}^{\text bev}, F_{\text rad}^{\text fv})$,
$ (F_{\text img}^{\text bev}, F_{\text rad}^{\text bev})$,
$ (F_{\text img}^{\text fv}, F_{\text rad}^{\text bev})$,
$ (F_{\text img}^{\text fv}, F_{\text rad}^{\text fv})$,
$ (F_{\text rad}^{\text bev}, F_{\text rad}^{\text fv})$.
Here, $ F^{\text bev} $ and $ F^{\text fv} $ represent bird’s-eye view and front view(FV) features, respectively, while  $F_{\text img}$ and $F_{\text rad}$ denote image and radar features, respectively. Before calculating the Global Contrastive Loss, features are further processed using an attention mechanism to compute weighted representations. For illustration, two feature representations, $ (F_{\text img}^{\text bev}, F_{\text img}^{\text fv})$ are used as an example, with each having dimensions ${C \times H \times W}$, where $H$ and $W$ represent the spatial feature height and width, and $C$ denotes the number of channels. 

To integrate the features from both modalities, we apply a weighted summation approach. Instead of relying on attention weights derived from a single modality, which could lead to the loss of crucial information from the other modality, we first concatenate the features. This concatenation enables the attention module to simultaneously consider the feature distributions from both views, ensuring a more balanced and comprehensive weight distribution.
For the first step of feature aggregation, shared attention weights are computed for the rows. These weights determine the importance of each row by comparing the contributions of both views in the concatenated feature representation. The row weights are then used to perform a weighted summation over the rows of the BEV and FV features, resulting in aggregated column-wise feature representations for each modality. This operation captures the key information across rows, producing features that are condensed along the row dimension.
Next, similar to row aggregation, we calculate attention weights for the columns using the previously aggregated features. These column weights are computed based on the concatenated column-wise features of the two modalities. The column weights are then used to perform a weighted summation across columns, yielding the final global feature representations for both modalities.
Through this two-step process of row-wise and column-wise aggregation, we achieve a global feature representation for each modality, $F_{\text {global}}^{bev}$ and $F_{\text {global}}^{fv}$. These global features are now in a compact $C$-dimensional form that better characterizes the overall scene.

For a given feature map, positive samples correspond to the features from the same batch that represent the same scene. For example,  for the contrastive loss $ \mathcal{L}(F_{\text {img}}^{\text {bev}}, F_{\text {img}}^{\text {fv}})$, the positive samples are the feature from the same batch, denoted as $F_{\text{img}}^{\text{bev}(i)}$ and $F_{\text{img}}^{\text{fv}(i)}$, where $i$ is the index of the scene within the batch. Negative samples are those features that belong to different scenes within the same batch.
The contrastive loss for $ \mathcal{L}(F_{\text {img}}^{\text {bev}}, F_{\text {img}}^{\text {fv}})$ can be formulated as follows:
\vspace{-7pt}
{
\footnotesize
\begin{equation}
\mathcal{L}\left( F_{\text{img}}^{\text{bev}}, F_{\text{img}}^{\text{fv}} \right) = 
-\log \frac{
    \exp \left( \operatorname{sim} \left( F_{\text{global}}^{\text{bev}(i)}, F_{\text{global}}^{\text{fv}(i)} \right) / \tau \right)
}{
    \sum_{k=1}^{N} \exp \left( \operatorname{sim} \left( F_{\text{global}}^{\text{bev}(i)}, F_{\text{global}}^{\text{fv}(k)} \right) / \tau \right)
}
\end{equation}
}

The overall Global Contrastive Loss is then computed as the sum of the individual contrastive losses for all the positive and negative sample pairs across all sets of features:

\vspace{-10pt}
{\footnotesize
\begin{equation}
\begin{aligned}
    \mathcal{L}_{\text{global}} =\;
     &\mathcal{L}\left( F_{\text{img}}^{\text{bev}}, F_{\text{img}}^{\text{fv}} \right)
    + \mathcal{L}\left( F_{\text{img}}^{\text{bev}}, F_{\text{rad}}^{\text{fv}} \right)
    + \mathcal{L}\left( F_{\text{img}}^{\text{fv}}, F_{\text{rad}}^{\text{fv}} \right) \\
    &+ \mathcal{L}\left( F_{\text{img}}^{\text{bev}}, F_{\text{rad}}^{\text{bev}} \right)
    + \mathcal{L}\left( F_{\text{img}}^{\text{fv}}, F_{\text{rad}}^{\text{bev}} \right)
    + \mathcal{L}\left( F_{\text{rad}}^{\text{bev}}, F_{\text{rad}}^{\text{fv}} \right)
\end{aligned}
\end{equation}
}

\noindent Finally, we experimentally verify that the weight of global contrastive loss is 1/6 of the best effect in the appendix, so our overall contrastive loss is:
${\mathcal L} = {\mathcal L}_{\text {global }} / 6 + {\mathcal L}_{\text {col }}$.

\section{Experiment}

\subsection{Implementation Details}


\noindent \textbf{Backbones:} 
Our experiments used CRN, RCBEVDet, and BEVFusion \cite{liu2023bevfusion} as backbones to evaluate our methods. CRN employs multi-modality deformable attention to transform image features into a semantically rich bird’s-eye-view (BEV) representation. BEVFusion, originally designed for lidar-camera fusion, was adapted for radar-camera fusion to enable direct comparison. RCBEVDet integrates RadarBEVNet and a Cross-Attention Multi-layer Fusion module to align and merge radar and camera data. All backbones were configured using the best publicly available settings from their open-source implementations.

\noindent \textbf{Dataset:} 
Experiments utilized the NuScenes \cite{caesar2020nuscenes} and Lyft Level 5 \cite{houston2021one} datasets, established benchmarks in autonomous driving. NuScenes provides 3D object annotations from a full sensor suite across diverse urban scenarios, while Lyft Level 5 includes over 1,000 hours of lidar and camera data tracking vehicles, cyclists, and pedestrians.

\noindent \textbf{Metrics:} To assess the performance of our methods, we used standard evaluation metrics, including mean Average Precision (mAP) and NuScenes Detection Score (NDS)\cite{caesar2020nuscenes}. These metrics provide a comprehensive assessment of detection accuracy and overall system performance.


\subsection{Main Results}

\begin{table}[h!]
\caption{\small{Comparison of overall 3D object detection performance on the NuScenes validation set. BEVFusion* represents a modified model for radar-image fusion. Metrics include mean Average Precision (mAP) and NuScenes Detection Score (NDS) with pretraining improvements shown in parentheses. \textbf{(w. CLLAP)} indicates results using our proposed CLLAP framework.}}
\centering
\setlength{\tabcolsep}{8pt} 
\vspace{-10pt}
\begin{adjustbox}{width=0.46\textwidth}
\begin{tabular}{@{}lcc@{}}
\toprule
\textbf{Method} & \textbf{mAP$\uparrow$} & \textbf{NDS$\uparrow$} \\
\midrule
CenterFusion \cite{nabati2021centerfusion} & 33.20 & 45.30 \\
CRAFT \cite{kim2023craft} & 41.10 & 51.70 \\
RCBEV4d \cite{zhou2023bridging} & 45.30 & 56.30 \\
SOLOFusion \cite{park2022time} & 42.70 & 53.40 \\
StreamPETR \cite{wang2023exploring} & 43.20 & 54.00 \\
\cmidrule{1-3}
BEVfusion* \cite{liu2023bevfusion} & 34.89 & 35.26 \\
BEVfusion* \textbf{(w. CLLAP)} & 38.41 (+3.52) & 38.20 (+2.94) \\
\cmidrule{1-3}
CRN \cite{kim2023crn} & 47.30 & 55.74 \\
CRN \textbf{(w. CLLAP)} & \textbf{50.53 (+3.23)} & \textbf{58.02 (+2.28)} \\
\cmidrule{1-3}
RCBEVDet \cite{lin2024rcbevdet} & 45.30 & 56.85 \\
RCBEVDet \textbf{(w. CLLAP)} & 45.91 (+0.61) & 57.09 (+0.24) \\
\bottomrule
\end{tabular}
\end{adjustbox}
\vspace{-13pt}

\label{tab:general_metrics}
\end{table}

\cref{tab:general_metrics} summarizes the performance gains brought by our pretraining on the nuScenes validation set. On CRN, CLLAP yields an absolute improvement of +3.23 mAP and +2.28 NDS. BEVFusion* also benefits notably, with mAP and NDS increased by +3.52 and +2.94, respectively. Even the strong RCBEVDet baseline is further improved, achieving gains of +0.61 mAP and +0.24 NDS. Compared with state-of-the-art radar–camera fusion methods, models equipped with our pretraining framework achieve superior overall performance, indicating that CLLAP is both effective and backbone-agnostic. To further assess the generalization ability of our pretraining, we also evaluate these models under adverse-weather conditions; detailed per-condition results are reported in the appendix.


\vspace{-5pt}
\subsection{L2R Sampling}
To evaluate the effectiveness of the proposed L2R Sampling module, we conducted experiments using VoxelNet \cite{zhou2018voxelnet} as the benchmark. The baseline used the NuScenes radar training dataset, while the ``sample by distance" setting employed pseudo-radar points sampled with distance-based weights. The L2R Sampling module generated pseudo-radar data using our proposed strategy, with the ``map" setting further projecting sampled points onto the radar plane to emulate radar's altitude insensitivity.

As shown in \cref{tab:L2R_Sampling_Module}, the L2R Sampling module consistently outperformed the baselines in mAP and NDS. (Results for different sampling methods are shown in the appendix.) These improvements stem from the module's ability to better preserve radar point cloud distributions and retain object-relevant points.
Additionally, to quantify its similarity to real radar data, we calculated the Chamfer Distance \cite{fan2017point}. Over 4000 frames, the L2R Sampling module achieved a mean Chamfer Distance of 112.16, compared to 120.1 for the "sample by distance" method, confirming its ability to better replicate radar point structures.

\begin{table}[!t]
\caption{\small{Comparison of L2R Sampling module 3D object detection performance on NuScenes val set.}}
\centering
\vspace{-10pt}

\resizebox{0.45\textwidth}{!}{%
\begin{tabular}{lccc}
\toprule
\textbf{Training Data} & \textbf{mAP} ($\uparrow$) & \textbf{NDS} ($\uparrow$)  \\ 
\midrule
Real radar data & 7.40 & 31.36  \\ 
\cmidrule{1-3}
Sample by distance & 11.25(+3.85) & 32.32(+0.96) \\ 
L2R Sampling module & \textbf{12.73(+5.33)} & \textbf{33.49(+2.13)}  \\ 
\cmidrule{1-3}
Sample by distance(map) & 10.57(+3.17) & 32.53(+1.17) \\ 
L2R Sampling module(map) & 11.02(+3.62) & 32.75(+1.39)  \\ 
\bottomrule
\end{tabular}%
}
\vspace{-10pt}

\label{tab:L2R_Sampling_Module}

\end{table}

\subsection{Ablation Study}
\label{sec4.4}
\begin{table}[!t]
\caption{\small{Ablation of the components of Dual-Stage Dual-Modality Contrastive Learning Strategy.}}
\centering
\vspace{-10pt}
\resizebox{0.45\textwidth}{!}{%
\begin{tabular}{c|cc|ccc}
\toprule
\textbf{Config} & \textbf{Local} & \textbf{Global} & \textbf{mAP}($\uparrow$) & \textbf{NDS}($\uparrow$)  \\ 
\midrule
baseline & - & - & 47.30 & 55.74 \\
1 & \checkmark & - & 49.72 (+2.42) & 57.73 (+1.99) \\
2 & - & \checkmark & 48.13 (+0.83) & 56.28 (+0.54)  \\
3 & \checkmark & \checkmark & \textbf{50.53 (+3.23)}   & \textbf{58.02 (+2.28)} \\
\bottomrule
\end{tabular}%
}
\vspace{-10pt}

\label{tab:ablation_rtiesults}

\end{table}




\begin{table}[t]
\caption{Impact of Global Contrastive families in $\mathcal{L}_{\text{global}}$.}
\vspace{-10pt}
\centering
\small
\setlength{\tabcolsep}{4pt}

\resizebox{0.95\linewidth}{!}{%
\begin{tabular}{c|c|cc} 
\toprule
\textbf{Config} & \textbf{Component} & {\textbf{mAP}\,$\uparrow$} & {\textbf{NDS}\,$\uparrow$} \\
\midrule
baseline & None                         & 47.30 & 55.74 \\
1        & Same-mod. Cross-view        & 50.11(+2.81) & 57.80(+2.06) \\
2        & Cross-mod. Same-view         & 49.81(+2.51) & 57.83(+2.09) \\
3        & Cross-mod. Cross-view        & 50.18(+2.88) & 58.11(+2.37) \\
4        & All (three families)         & \textbf{50.53(+3.23)} & \textbf{58.02(+2.28)} \\
\bottomrule
\end{tabular}
}
\vspace{-10pt}
\label{tab:abl_local_losses}
\end{table}

\begin{table}[!t]
\caption{\small{Comparison of 3D object detection performance across different datasets pretrain on NuScenes val datasets. }}
\centering

\vspace{-10pt}
\resizebox{0.45\textwidth}{!}{%
\begin{tabular}{c|cc|ccc}
\toprule
\textbf{Config} & \textbf{Lyft} & \textbf{NuScenes} & \textbf{mAP}($\uparrow$) & \textbf{NDS}($\uparrow$)  \\ 
\midrule
baseline & -  & -  & 47.30        & 55.74                 \\
1 & \checkmark  & -  & 48.84 (+1.54)   & 56.87 (+1.13)      \\
2 & -  & \checkmark  & 48.60 (+1.30)   & 56.37 (+0.63)      \\
3 & \checkmark  & \checkmark  & \textbf{50.53 (+3.23)}   & \textbf{58.02 (+2.28)}      \\
\bottomrule
\end{tabular}%
}
\vspace{-15pt}

\label{tab:comparison_datasets}
\end{table}


We perform ablation studies on the NuScenes validation set using CRN as the backbone to evaluate the effectiveness of each setting of our proposed framework.

\noindent \textbf{Impact of Contrastive Losses.}
Our method leverages two contrastive loss components during pretraining to enhance model performance. As shown in \cref{tab:ablation_rtiesults}, each contrastive loss significantly improves model performance, with the combined losses yielding the highest gains. These results validate the effectiveness of both loss components in our contrastive learning framework.

\noindent
\noindent \textbf{Impact of Global Contrastive Losses.}
We decompose the global contrastive loss into three families and train three variants, each activating only one family under identical settings; \cref{tab:abl_local_losses} shows that all three consistently improve over the baseline. \emph{Same-modality cross-view} improves view invariance and geometric consistency across viewpoints. \emph{Cross-modality same-view} reduces the semantic gap between sensors under a shared projection, enabling smoother and better-calibrated fusion. \emph{Cross-modality cross-view} enforces compositional alignment under joint changes of viewpoint and modality, promoting cycle-consistent features and mitigating parallax- and occlusion-induced discrepancies. Overall, each family provides a distinct yet complementary benefit, jointly validating $\mathcal{L}_{\text{global}}$ as a principled composition of global contrastive constraints.


\noindent \textbf{Impact of Pretraining Datasets.}
To assess the impact of pretraining data volume, we conducted an ablation experiment, with results summarized in \cref{tab:comparison_datasets}. Pretraining on either Lyft Level 5 or nuScenes improves the downstream performance on the nuScenes validation set, while Lyft-only pretraining yields slightly larger gains than nuScenes-only pretraining. Combining both datasets leads to the best performance.

\begin{table}[!t]
\centering
\caption{\small{Ablation of the step components.}}
\vspace{-10pt}
\resizebox{0.46\textwidth}{!}{%
\begin{tabular}{c|ccc|ccc}
\toprule
\textbf{Config} & \textbf{Step 1} & \textbf{Step 2} & \textbf{Step 3} & \textbf{mAP}($\uparrow$) & \textbf{NDS}($\uparrow$)  \\ 
\midrule
1                       & -        & -        & \checkmark        & 47.30        & 55.74                    \\
2       & \checkmark        & -        & \checkmark        & 49.53 (+2.23)   & 57.77 (+2.03)       \\
3       & -        & \checkmark        & \checkmark        & 49.11 (+1.81)   & 57.57 (+1.83)        \\
4 & \checkmark        & \checkmark        & \checkmark        & \textbf{50.53 (+3.23)}   & \textbf{58.02 (+2.28)}        \\
\bottomrule
\end{tabular}%
}

\vspace{-15pt}
\label{Stage Components}

\end{table}
\noindent \textbf{Impact of the BCSA Module.} 
To isolate the benefit of BCSA beyond generic cross-attention, we replace it with a bi-directional cross-attention that applies a single multi-head attention along the spatial sequence, without channel–spatial decoupling or gated fusion. Under identical training, this vanilla variant yields 49.77 mAP / 57.57 NDS—only +0.08 mAP over removing the module entirely (49.69 / 57.74) while reducing NDS by 0.17—suggesting that naive cross-modal token mixing introduces noisy associations that do not improve overall driving quality. In contrast, BCSA attains 50.53 mAP / 58.02 NDS, indicating that its structured channel–spatial interactions and normalized, gated bi-directional fusion better exploit complementary cross-modal cues while suppressing spurious correlations, thereby improving both detection accuracy and the holistic NDS metric.

\noindent \textbf{Impact of the Framework Steps.}
We evaluated the contribution of each pertaining step in our framework, as summarized in \cref{Stage Components}. 
Step 1, which involves pre-training on pseudo-radar point clouds, improves mAP and NDS. This indicates that the L2R Sampling module and contrastive learning enhance generalization.
Step 2, using a small set of real radar data for secondary pre-training, further improves mAP and NDS, highlighting the value of real radar data for refining model accuracy. The best performance was achieved by combining all three steps, resulting in significant improvements across mAP and NDS. This demonstrates that integrating pseudo-radar data, real radar data, and fine-tuning yields the best performance.

\vspace{-10pt}
\section{Conclusion}
In this paper, we propose CLLAP, a novel Contrastive Learning-based LiDAR-Augmented Pretraining framework designed to enhance the performance and robustness of existing radar-camera fusion methods. We introduce the L2R module, which generates high-quality pseudo-radar data from large-scale LiDAR datasets, enhancing data diversity and facilitating robust model training. Additionally, we present a Dual-Stage Dual-modality Contrastive Learning Strategy that leverages features from different modalities and views to enhance the model's representation capabilities.
Extensive experiments on multiple datasets demonstrate the effectiveness of CLLAP and the superiority of its components.

\section{Acknowledgement}
This work was supported in part by the National Science and Technology Major Project under Grant 2024ZD1600100, in part by the National Natural Science Foundation of China under Grants 62272357 and 62302326, in part by the Wuhan Science and Technology Joint Project for Building a Strong Transportation Country under Grant 2024-2-7, in part by the Wuhan Science and Technology Project for Key Research and Development under Grant 2024050702030090, and in part by the Fundamental Research Funds for the Central Universities under Grant 2042025kf0055.

{
    \small
    \bibliographystyle{ieeenat_fullname}
    \bibliography{main}
}

\clearpage
\setcounter{page}{1}
\maketitlesupplementary
\setcounter{section}{0}

\section{Overview}
The appendix offers comprehensive explanations of the methodologies introduced in the main text, together with additional experimental results and extended visual analyses. The supplementary material is organized into the following
sections:
\begin{itemize}
\item Sec.\ref{sec2} Methodology Supplement
\begin{itemize}
\item Sec.\ref{sec2.1} Sliding Window Feature Matching Mechanism
\item Sec.\ref{sec2.2} BCSA Module
\item Sec.\ref{sec2.3} Column Selection in Local Contrastive Loss
\end{itemize}
\item Sec.\ref{sec3} Visual supplementation
\begin{itemize}
    \item Sec.\ref{sec3.1} Visualization of Experimental Results
    \item Sec.\ref{sec3.2} Global Contrastive Loss
\end{itemize}
\item Sec.\ref{sec4} Supplementary Experiments
\begin{itemize}
    \item Sec.\ref{sec adverse weather} Generalization to adverse weather
    \item Sec.\ref{sec4.1} Impact of the weighting parameter in loss function
    \item Sec.\ref{sec4.2} Impact of the GMM parameter variations
    \item Sec.\ref{sec4.3} Impact of the weighting parameter in sampling
    \item Sec.\ref{sec4.5} Comparison with Learned Pseudo-Radar Baselines
\end{itemize}
\end{itemize}

\section{Methodology Supplement}
\label{sec2}

\subsection{Sliding Window Feature Matching Mechanism}
\label{sec2.1}
Cross-modality feature misalignment presents a significant challenge in multi-modal contrastive learning for radar-camera fusion, as naively treating spatially corresponding features as positive pairs often results in suboptimal alignment. To address this limitation, we proposed a method for finding the best positive sample pairs. Specifically, for each anchor $\mathbf{f}_t^j$, We define a local search area with a width of $R$ in the query feature map and use a sliding window of size $r$ ($r$\textless$R$) to slide through the local search area to extract $n$ ($n=R-r+1$) candidate areas $\mathbf{f}_s^m$. Each candidate area is adaptively aggregated through an attention module $Atten$ to align with the anchor’s dimensions, followed by pairwise similarity computation. The highest-scoring candidate $\mathbf{f}_s^{j*}$ is selected as the refined positive pair, effectively replacing coarse positional alignment with a learned, similarity-driven matching strategy. This approach enhances feature correspondence precision while maintaining robustness to spatial discrepancies between pseudo-radar and camera modalities:
{
\begin{equation}
\begin{split}
\mathbf{f}_s^{j*} =
\underset{\delta \in \{-R, \ldots, R\}}
{\mathrm{max}}
&\biggl( \mathrm{sim} \left( \mathbf{f}_t^j, Atten \left( \vphantom{\Big|} \right. \right. \\
& \left. \left. \mkern-9mu \left\{ \mathbf{f}_s^m \mid \left| m - (j + \delta) \right| \leq \frac{r}{2} \right\} \right) \right) \biggr)
\end{split}
\end{equation}}

where $\operatorname{sim}(\cdot, \cdot)$ denotes similarity, $m$ is the center of the ${m}^{th}$ area during the sliding process, and $ \delta $ is the offset.

\subsection{BCSA Module}
\label{sec2.2}
To further enhance the discriminative power of these refined positive pairs, we introduce a Bidirectional Channel-Spatial Attention module. Within the constructed positive pairs, the channel-wise and spatial distribution information of the feature maps convey distinct semantic meanings. Our module independently applies attention mechanisms along the channel and spatial dimensions: 

{
\begin{equation}
\text{MAT}(Q,K,V)=\mathrm{softmax}\left(\frac{QK^\top}{\sqrt{d_k}}\right)V
\end{equation}}
where $Q=\mathbf{F}_i\in\mathbf{R}^{B\times D\times C}$ or $\Gamma(\mathbf{F}_i)\in\mathbf{R}^{B\times C\times D}$ is the query, $K, V=\mathbf{F}_{3-i}\in\mathbf{R}^{B\times D\times C}$ or $\Gamma(\mathbf{F}_{3-i})\in\mathbf{R}^{B\times C\times D}$ , and $D,C$ represent the spatial and channel dimensions, respectively.
(1) \textit{Channel Attention} learns importance weights for different feature channels, highlighting those most discriminative for the 3D object detection task. (2) \textit{Spatial Attention} learns importance weights for different spatial locations within the feature column, emphasizing the importance of the region of interest in object detection while suppressing interference from background regions, thus enhancing the alignment of pseudo-radar and camera features. Through this bidirectional decoupling and weighting across channel and spatial dimensions, the final refined feature representation is computed as:

\begin{figure}[!h]
\centering
\includegraphics[width=0.46\textwidth]{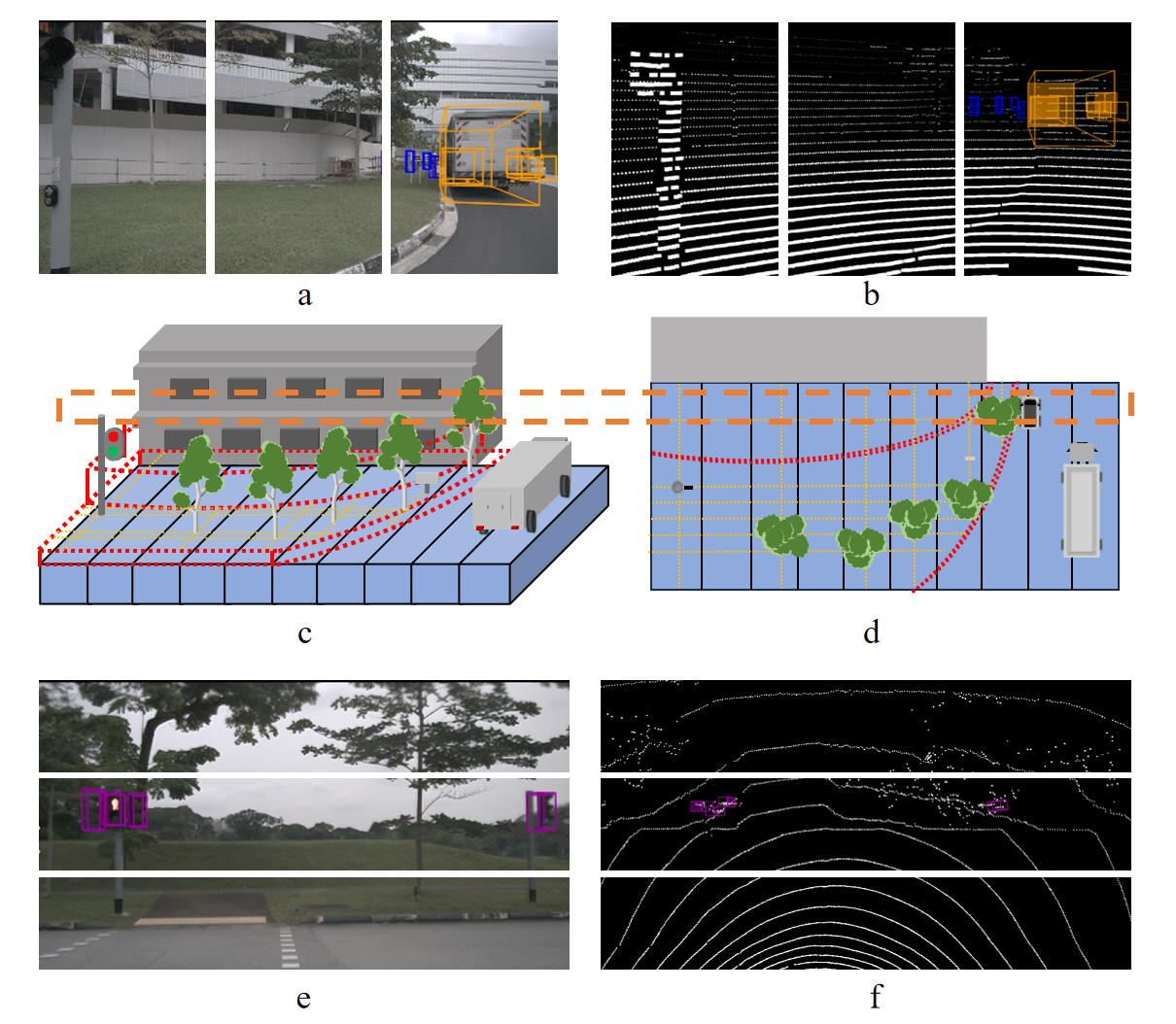}
\vspace{-10pt}
\caption{Figures (a) and (b) present the visualizations of the image and point cloud from the front view, respectively. Figures (c) and (d) illustrate simulated representations of the same modality observed from different viewpoints. Figures (e) and (f) depict visualizations of different modalities under varying viewpoints. 
}
\label{difference}
\end{figure}

{
\begin{equation}
\begin{split}
&\mathrm{F}_i^{\mathrm{final}} =\mathcal{G} \biggl( 
  \mathrm{LN}\left( \mathrm{MAT}(\mathbf{F}_i, \mathbf{F}_{3-i}, \mathbf{F}_{3-i}) \right), \\
 & \mathrm{LN}\left( \Gamma\left( \mathrm{MAT}(\Gamma(\mathbf{F}_i), \Gamma(\mathbf{F}_{3-i}), \Gamma(\mathbf{F}_{3-i})) \right) \right) 
\biggr)
\end{split}
\end{equation}
}
where $i \in \{1,2\}$ indexes the feature pair, $\mathcal{G}$ is a gate control mechanism, LN denotes layer normalization, and $\Gamma$ represents the transposition operation.

The following is the explanation of why columns are selected instead of rows in the local contrastive loss part of the two-stage dual-modal comparative learning strategy in the proposed method paragraph.


\subsection{Column Selection in Local Contrastive Loss}
\label{sec2.3}
As illustrated in \cref{difference}, Figures (a) and (b) indicate that both modalities convey consistent content when evaluated within the same column unit. Figures (c) and (d) demonstrate that column-based units maintain consistent content representation across views, whereas row-based units introduce noticeable inconsistencies, as highlighted by the yellow dashed rectangles. Figures (e) and (f) reveal that row-based optimization units exacerbate inconsistencies in content representation across modalities and views.
\section{Visual supplementation}
\label{sec3}
\subsection{Visualization of Experimental Results}
\label{sec3.1}

\begin{figure}[!htbp]
\centering
\includegraphics[width=0.45\textwidth]{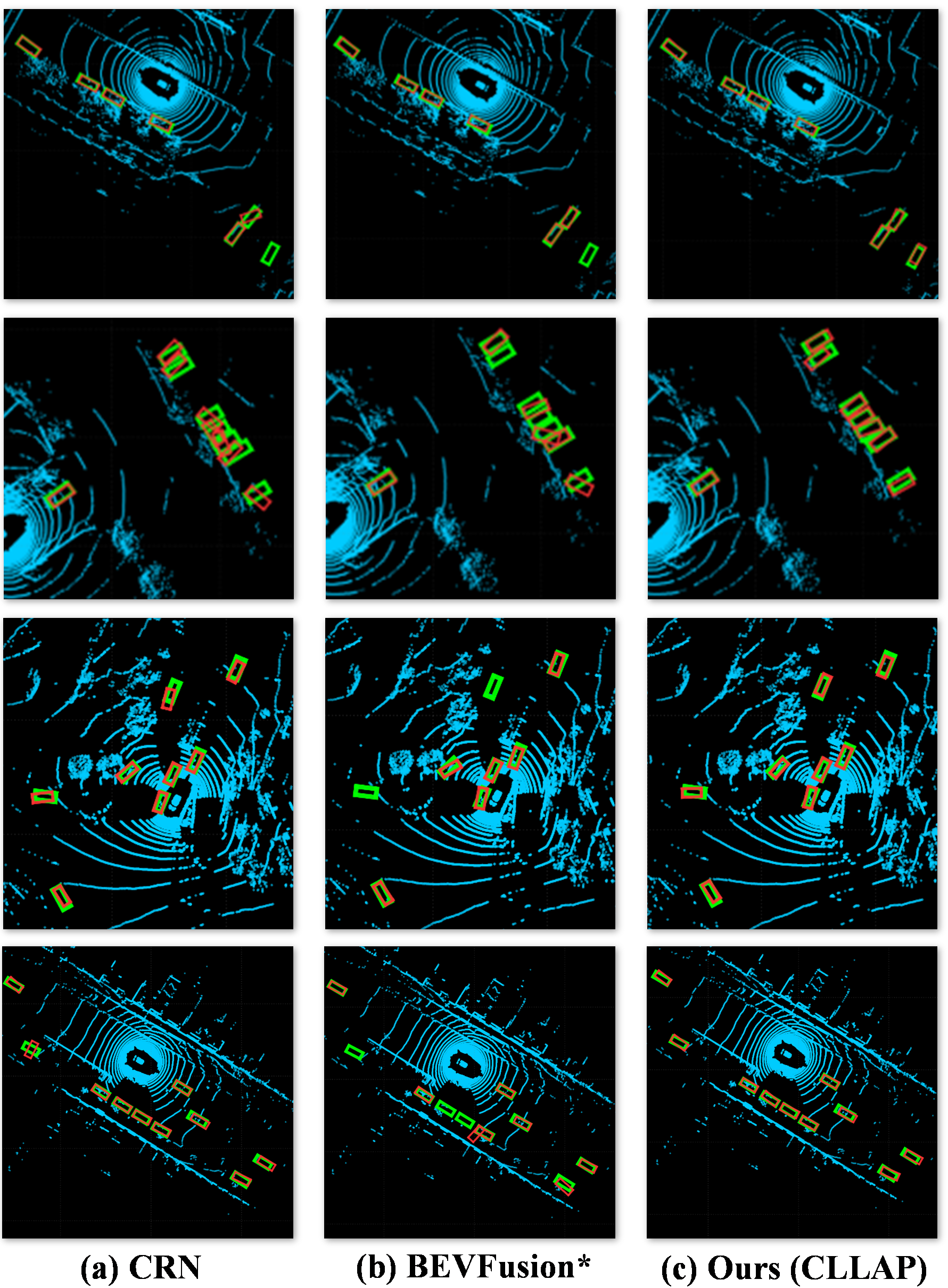}
\vspace{-10pt}
\caption{Comparison of our method with baseline visualization results.}
\label{show}
\end{figure}
Figure \ref{show} provides a visual comparison between the results produced by our proposed method and those generated by the CRN baseline. The green solid rectangle denotes the ground truth bounding box, the red dotted rectangle represents the prediction from the baseline model, and the blue dotted rectangle corresponds to the prediction obtained using our method. As observed, the predictions produced by our approach exhibit greater alignment with the ground truth, demonstrating improved localization accuracy compared to the baseline. This visual evidence highlights the effectiveness of our method in enhancing detection performance.

\subsection{Global Contrastive Loss}
\label{sec3.2}
\begin{figure}[!h]
\centering
\includegraphics[width=0.46\textwidth]{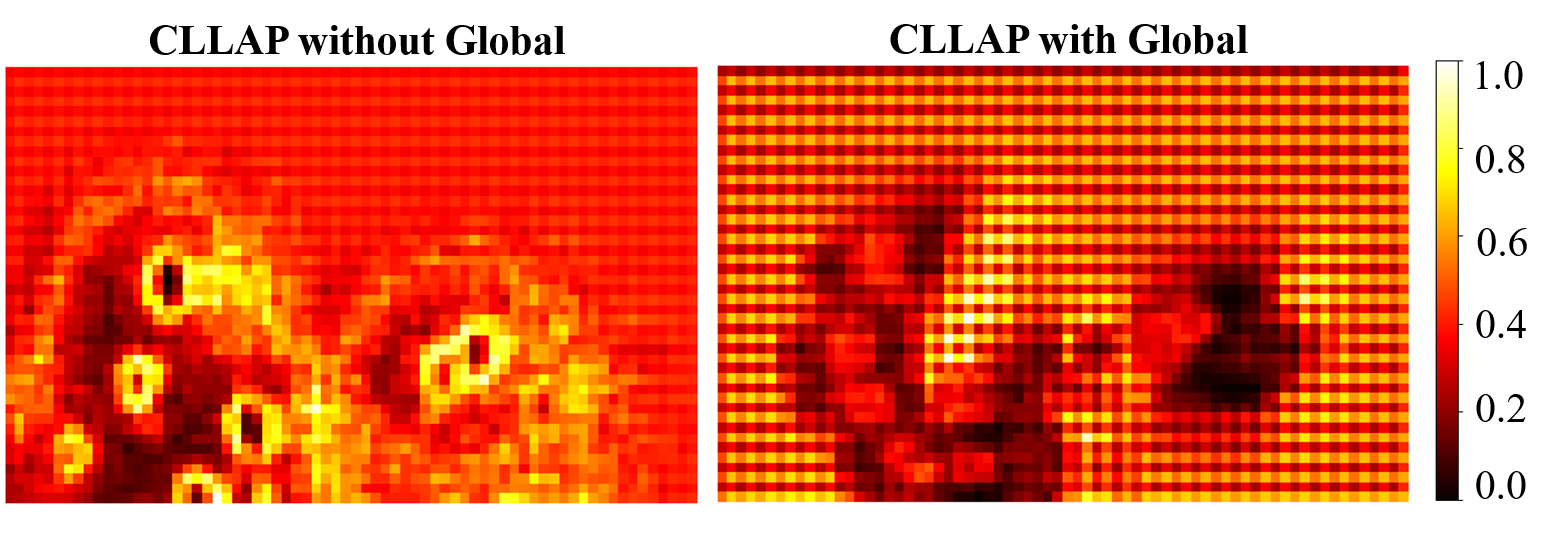}
\vspace{-10pt}
\caption{Comparison of feature heatmaps before and after the application of Global Contrastive Loss is presented.}
\label{global}
\end{figure}
As illustrated in \cref{global}, the left panel shows the inconsistency map produced by CLLAP without the Global Contrastive Loss, while the right panel shows the result after adding this loss. In both maps, brighter colors indicate a larger discrepancy between each feature and its local neighborhood. Without the global supervision, high-response regions are scattered across both foreground objects and background areas, and even the interior of the target exhibits many fragmented, high-valued blobs, suggesting unstable and noisy representations. After introducing the Global Contrastive Loss, the responses become much more compact: large values concentrate around object boundaries, whereas the interior of the object and most background regions are strongly suppressed. This indicates that the proposed loss encourages globally coherent, object-level features and clearer separation between foreground and background, which aligns with the quantitative gains in detection accuracy reported in \cref{sec4.4}. 



\section{Supplementary Experiments}
\label{sec4}

\noindent \textbf{Implementation Settings.}
{Our proposed model is implemented using the \textit{PyTorch} framework and trained on \textit{NVIDIA GeForce RTX 4090} and \textit{NVIDIA H800 Tensor Core GPUs}. 
We adopt the \textit{SGD} optimizer with a learning rate of \(2\times10^{-4}\), momentum of \(0.9\), and weight decay of \(1\times10^{-5}\). 
The batch size is set to \(6\) during pretraining.}


\begin{figure}[!h]
\centering
\includegraphics[width=0.46\textwidth]{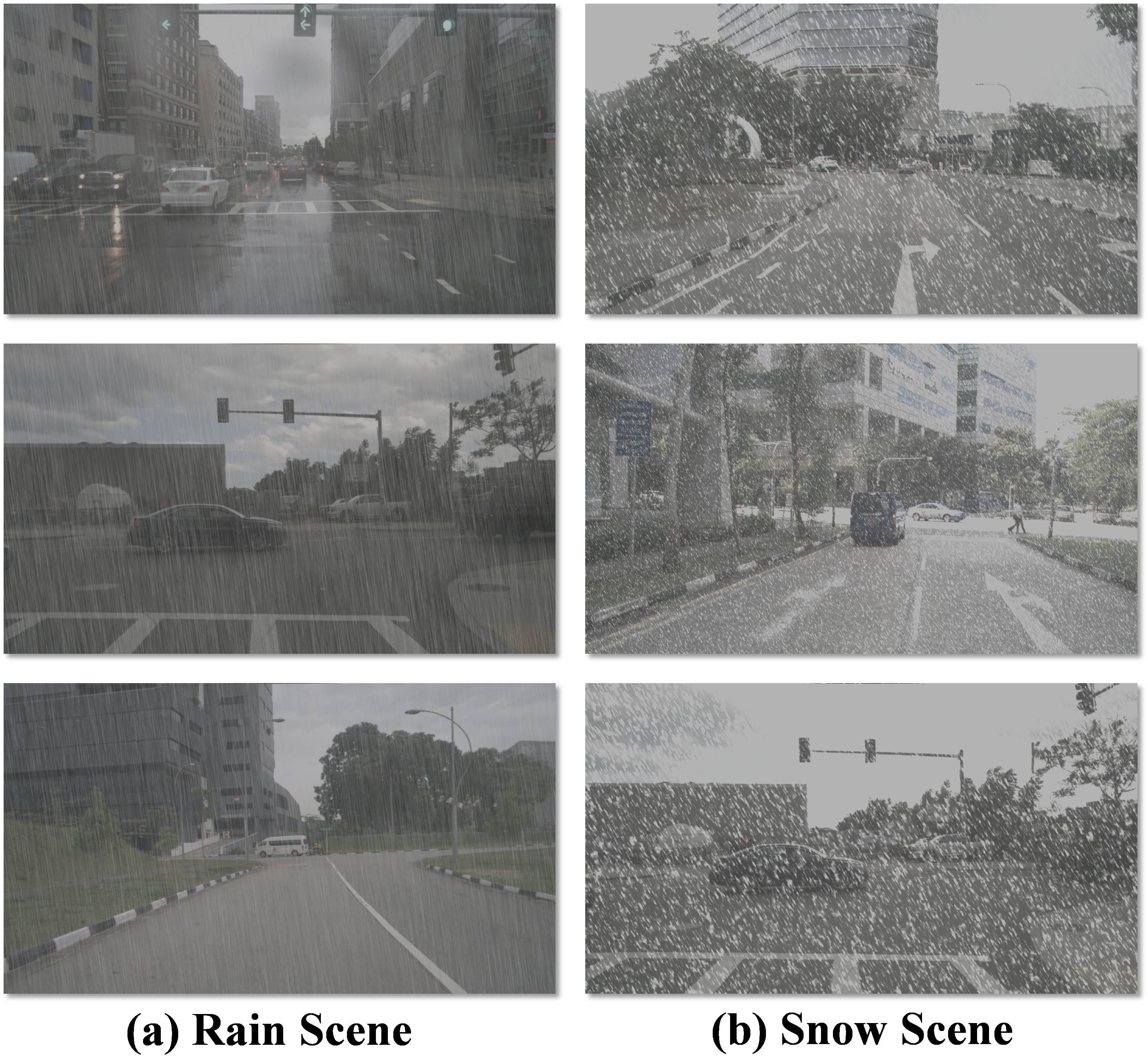}
\caption{Adverse Weather example}
\label{fig:adverse_weather_visual}
\end{figure}

\subsection{Generalization to adverse weather}
\label{sec adverse weather}
To further assess whether the proposed pretraining improves the robustness of the learned representations, we evaluate the models on adverse-weather scenarios without any additional training. Following the corruption protocol of the CVPR 2023 work\cite{dong2023benchmarking}, we synthesize adverse-weather data by injecting noise and weather-related corruptions into the original nuScenes samples; representative examples are shown in \cref{fig:adverse_weather_visual}. We then directly apply the models pretrained with our method and the corresponding baselines to this corrupted benchmark, using exactly the same weights as on the clean nuScenes validation set. As summarized in \cref{tab:adverse_weather}, our pretrained models consistently outperform their non-pretrained counterparts under adverse-weather conditions, suggesting that the learned features generalize beyond the clean training distribution and confer improved robustness to realistic corruptions.

\subsection{Impact of the weighting parameter in the loss function}
\label{sec4.1}
To study how the balance between the global and column-wise contrastive terms affects performance, we ablate the weighting factor $\lambda$ in the loss:
{

\begin{equation}
    \mathcal{L} = \lambda \mathcal{L}_{\text{global}} + \mathcal{L}_{\text{col}}.
\end{equation}}
For efficiency, all models are trained on a randomly sampled 20\% subset of the nuScenes training set. As reported in \cref{tab:loss_weight}, we evaluate $\lambda \in \{1, 1/6, 1/12\}$. Setting $\lambda = 1/6$ yields the best detection accuracy, indicating that a moderate weight on the global term provides a good trade-off between global regularization and the local column-wise supervision. In contrast, $\lambda = 1$ over-emphasizes the global loss and slightly degrades performance, while $\lambda = 1/12$ under-weights it and limits the benefit of global contrastive learning. Therefore, we adopt $\lambda = 1/6$ as the default in all experiments.

\subsection{Impact of the GMM parameter variations}
\label{sec4.2}
Additionally, to verify that GMM parameter variations have a minimal impact on our framework, we evaluated GMM with 4, 5, and 6 components by Chamfer distances(a measure of similarity between two sets):
{
\begin{equation}
\begin{split}
&CD(P, Q) = \\
&\frac{1}{|P|} \sum_{p \in P} \min_{q \in Q} | p - q |^2 + \frac{1}{|Q|} \sum_{q \in Q} \min_{p \in P} | p - q |^2
\end{split}
\end{equation}}
where P/Q denote the pseudo-radar/real-radar points.
The results shown in \cref{tab:GMM settings} reveal that the setting of GMM parameters had little effect on the results.

\subsection{Impact of the weighting parameter in sampling}
\label{sec4.3}
In the paper, the intensity weight $w_{\mathrm{int}}$ reflects the relative contribution of each point based on its point intensity. To ensure points with greater reflective intensity are properly emphasized during sampling, greater weights are assigned to them. The intensity weight for each point is calculated as $w_{\mathrm{int}} = I_i^{\frac{1}{2}} / \sum_j I_j^{\frac{1}{2}}$, where $I_i$ is the intensity of point $i$. 


\begin{table}[t]
    \centering
    \caption{\small{Performance under adverse-weather corruptions on the nuScenes validation set. Models are directly evaluated on corrupted data without additional training.}}
    \label{tab:adverse_weather}
    \scalebox{0.95}{
    \begin{tabular}{lcccc}
        \toprule
        Weather & \multicolumn{2}{c}{Baseline} & \multicolumn{2}{c}{w.\ CLLAP (ours)} \\
        \cmidrule(lr){2-3} \cmidrule(lr){4-5}
        & mAP ($\uparrow$) & NDS ($\uparrow$) & mAP ($\uparrow$) & NDS ($\uparrow$) \\
        \midrule
        Rain & 32.84 & 46.57 & \bf{34.39} & \bf{47.09} \\
        Snow & 13.93 & 33.40 & \bf{15.45} & \bf{34.48} \\
        \bottomrule
    \end{tabular}
    }
\end{table}

\begin{table}[!t]
\centering
\caption{The result of different weight settings in the loss factor (using 1/5 of the nuScenes dataset).}
\label{tab:loss_weight}
\scalebox{0.95}{
\begin{tabular}{cc}
\toprule
\textbf{weight} & \textbf{mAP}($\uparrow$) \\ 
\midrule
1      & 35.13 \\
1/6    & \textbf{35.74} \\
1/12   & 35.29 \\
\bottomrule
\end{tabular}
}
\end{table}

\begin{table}[!t]
\caption{The result of different component settings in GMM.}
\centering
\scalebox{0.95}{
\begin{tabular}{c|c}
\toprule
\textbf{components} & \textbf{Chamfer Distances}($\downarrow$) \\ 
\midrule
4 & 142.0 \\
5 & 140.7 \\
6 & 141.2  \\
\bottomrule
\end{tabular}
}

\label{tab:GMM settings}
\end{table}

\begin{table}[!t]
\centering
\caption{The result of different sampling weight configurations.}
\scalebox{0.95}{
\begin{tabular}{c|ccc|c}
\toprule
\textbf{Config} 
& $\boldsymbol{\alpha_{\text{int}}}$ 
& $\boldsymbol{\alpha_{\text{dist}}}$ 
& $\boldsymbol{\alpha_{\text{spa}}}$ 
& \textbf{Chamfer Distances}($\downarrow$) \\
\midrule
1 & 4 & 2 & 4 & \textbf{114.1} \\
2 & 1 & 1 & 1 & 116.6 \\
3 & 1 & 2 & 1 & 118.6 \\
4 & 2 & 1 & 1 & 115.3 \\
5 & 1 & 1 & 2 & 115.5 \\
\bottomrule
\end{tabular}
}
\label{tab:sampling_weights}
\end{table}

To ensure accurate feature representation, we assign weights based on point sparsity and distance. Sparse regions require greater sampling to capture their structural and semantic information, quantified by the sparsity weight $w_{\mathrm{spa}}={\sum_jD_{ij}^2}$, where $D_{ij}$ is the Euclidean distance between point $i$ and its $j$-th nearest neighbor. Similarly, points farther from the center of the point cloud, often sparse in radar sensing, are assigned a distance weight  $w_{\mathrm{dist}}={1} /{D_{iO}^2}$, where $D_{iO}$ is the Euclidean distance from point $i$ to the origin. 
Finally, the overall sampling weight is a linear combination of the three individual weights, with scaling factors $\alpha_{int}$, $\alpha_{spa}$, $\alpha_{dist}$ controlling the contributions of each weight:
$w_{\mathrm{final}}=\alpha_{\mathrm{int}}w_{\mathrm{int}}+\alpha_{\mathrm{dist}}w_{\mathrm{dist}}+\alpha_{\mathrm{spa}}w_{\mathrm{spa}}$. By normalizing $w_{\mathrm {final}}$ for all points, we ensure the sampling probabilities are well-defined. Furthermore, we conducted a sensitivity analysis of the weight using 2000 nuScenes point clouds with 5 settings: 4:2:4, 1:1:1, 1:2:1, 2:1:1, and 1:1:2. The results in \cref{tab:sampling_weights} show that the chosen weight of 4:2:4 offers the best performance and the model is relatively robust to weight variations. 

\subsection{Comparison with Learned Pseudo-Radar Baselines}
\label{sec4.5}
\begin{figure}[!t]
\centering
\includegraphics[width=0.45\textwidth]{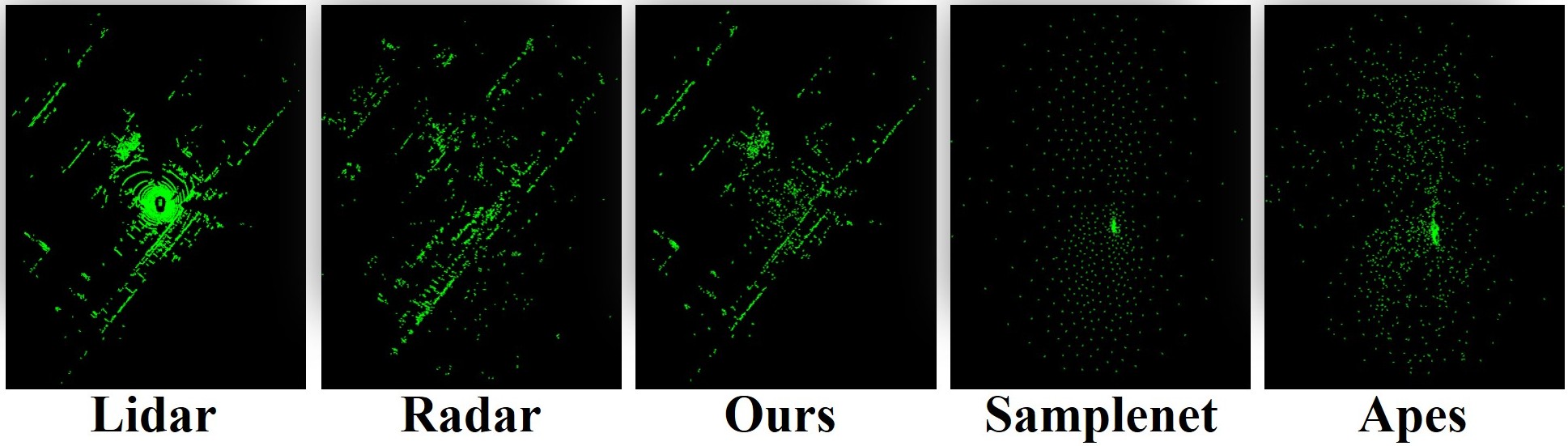}
\caption{\small Visualization Comparison}
\label{rebuttal_fig}
\end{figure}
We further compare the proposed L2R sampling strategy with two learning-based pseudo-radar generation baselines, namely SampleNet\cite{lang2020samplenet}(\emph{CVPR 2020}) and APES\cite{wu2023attention} (\emph{CVPR 2023}). As shown in Fig.~\ref{rebuttal_fig}, although these methods achieve lower Chamfer Distance values (SampleNet: \texttt{20.9}, APES: \texttt{42.0}), their generated point patterns are less consistent with the spatial characteristics of real radar observations. In contrast, the pseudo-radar point clouds produced by our L2R sampling strategy more faithfully preserve radar-like spatial distributions.

To further evaluate their utility for downstream pretraining, we incorporate SampleNet-generated pseudo-radar point clouds into the same pretraining pipeline used in our method. This variant improves the baseline by \texttt{+1.59} mAP and \texttt{+1.32} NDS, but still remains clearly inferior to our L2R-based pretraining, which achieves \texttt{+3.20} mAP and \texttt{+1.87} NDS. These results suggest that a lower Chamfer Distance alone does not necessarily translate into more effective pseudo-radar supervision for radar-camera fusion pretraining. More importantly, they demonstrate that the proposed L2R sampling strategy provides pseudo-radar data that are more beneficial for downstream 3D object detection.

\end{document}